\documentclass[runningheads]{llncs}

\usepackage{eccv}

\usepackage{eccvabbrv}

\usepackage{graphicx}
\usepackage{booktabs}
\usepackage{enumitem}
\setlist[itemize]{noitemsep, topsep=2pt, leftmargin=*}

\usepackage[accsupp]{axessibility}  %

\usepackage{hyperref}

\usepackage{orcidlink}

\begin{document}

\title{Authoring for Living Worlds: Tool-Constrained LLM Agents for Executable Multi-Actor Scenarios}

\titlerunning{Authoring for Living Worlds}

\author{Nicolae Cudlenco\inst{1,4}\orcidlink{0000-0001-6547-3659} \and
Mihai Masala\inst{2}\orcidlink{0000-0003-3496-9058} \and
Marius Leordeanu\inst{1,2,3}\orcidlink{0000-0001-8479-8758}}
\authorrunning{N.~Cudlenco et al.}
\institute{Institute of Mathematics of the Romanian Academy, Bucharest, Romania\\
\email{\{nicolae.cudlenco,leordeanu\}@gmail.com} \and
National University of Science and Technology Politehnica Bucharest, Romania\\
\email{mihaimasala@gmail.com} \and
Norwegian Research Center (NORCE), Bergen, Norway\\
\url{norceresearch.no} \and
B\"uchi Labortechnik AG, Flawil, Switzerland\\
\email{cudlenco.n@buchi.com}}

\maketitle

\begin{abstract}
Authoring a multi-actor scenario for a living 3D world, where every action changes its state, and each action's validity depends on the state accumulated before it, demands the freedom of storytelling and the rigor of simulation at once. We author such scenarios with LLM agents, as Graphs of Events in Space and Time (GESTs) that a simulation engine executes deterministically into narrative videos with per-frame spatial, temporal, and semantic ground truth. A staged pipeline driving a flagship LLM, the standard design in video generation, failed outright: the model violates rules stated verbatim in its prompt, and cannot track the dynamic world state. We answer with a constraint-enforcing tool layer: our Director and Scene Builder agents explore the world's capabilities page by page and build every scene through operations checked against simulator state, so every specification they emit is valid by construction. Because we generate each seed text from an existing scenario graph, we can measure reconstruction: the agent authors its own graph from the text alone, yet matches the original at 0.83 F1 on its events, each action with its participants (0.55 for a random scenario of the same kind), and 0.77 on their ordering (0.43 random). End to end: the standard staged pipeline produced 0 executable specifications in 50 attempts; our agents, driving a budget model, execute 20 of 25 (80\%), and are, to our knowledge, the first to exercise the full expressive capacity of GEST.
\end{abstract}

\section{Introduction and Related Work}
\label{sec:intro}

\begin{figure}[t]
\centering
\includegraphics[width=\textwidth]{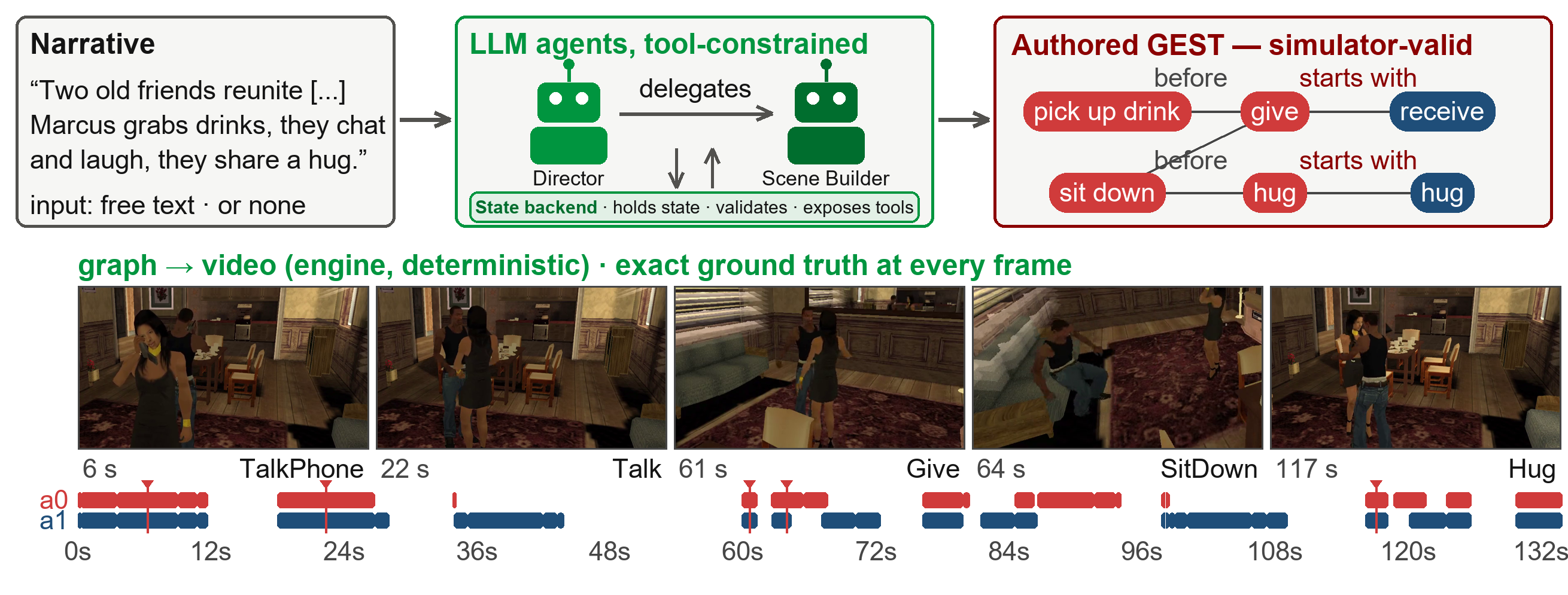}
\caption{\textbf{From narrative to executable world events.} An LLM Director and Scene Builder author a GEST through a constraint-enforcing tool layer: the world returns the actions it affords and rejects invalid operations, so the specification is valid by construction, including synchronizing interactions through complex temporal relations. The engine executes the graph deterministically, mapping events to frames (bottom).}
\label{fig:hero}
\end{figure}

A living world is only as alive as what can be authored for it. This paper starts from an explicit world model, the GEST-Engine~\cite{cudlenco2026tiny,cudlenco2026gestengineeventgraphssynthetic}. Its input is a Graph of Events in Space and Time (GEST)~\cite{masala2023gest}: a directed graph whose nodes are events (actions with multiple entities, performed by actors in specific locations) and whose edges carry temporal constraints from Allen's interval algebra~\cite{allen1983maintaining}, with support for logical and semantic relation types (Fig.~\ref{fig:hero}). The engine verifies the graph, then executes it deterministically, through the Multi Theft Auto framework, in GTA San Andreas: a shipped game world with 70+ environments, 733 objects, 2,500+ animations, and 312 character skins. The output is a film plus the world behind it: two to six synchronized actors play a multi-scene story under automatic camera tracking, and the engine records per-frame entity and camera state, all pairwise spatial relations, instance segmentation, exact event-to-frame boundaries, and a graph-derived description~\cite{cudlenco2026gtasagroundtruthannotations,cudlenco2026gestengineeventgraphssynthetic}. These videos hold up to human judgment: on the corpus GTASA~\cite{cudlenco2026gtasagroundtruthannotations}, 16 annotators found 69\% physically valid (VEO~3.1: 18\%, WAN~2.2: 13\%) and closer to the prescribed story (semantic match 4.09/5 versus 2.50 and 1.75). But so far the engine has run randomly generated graphs. We introduce the authoring: LLM agents turn free text, given or conceived through their own exploration of the world, into the complete executable specification: the cast, the locations, and every actor's chain of actions with its synchronizations and temporal constraints, in a world whose state every action changes.

LLM agents for video generation exist. StoryAgent~\cite{hu2024storyagent}, MAViS~\cite{wang2026mavis}, and DreamRunner~\cite{wang2026dreamrunner} orchestrate specialized agents across scriptwriting, casting, and rendering; VideoDirectorGPT~\cite{lin2023videodirectorgpt} and Dysen-VDM~\cite{fei2024dysen} condition diffusion on LLM-built plans. Every one of these pipelines ends the same way: a neural generator gets the final word, objects appear and disappear, actors morph, temporal order breaks~\cite{brooks2024sora,wan2025wan}, and since no plan-level artifact survives generation, there is nothing against which the output could be checked. None reports an executability figure; in their setting the concept does not exist.

Agents have driven engines before. GPT4Motion~\cite{lv2024gpt4motion} writes Blender scripts for physics-guided generation, but commands three basic motion primitives. FilmAgent~\cite{xu2025filmagent} stages multi-agent film production in Unity, choosing from predefined stage positions, actions, and camera shots. Its world is a menu: enumerable, stateless, every selection valid, so executability is not a question one can even ask there. The concurrent Cutscene Agent~\cite{he2026cutscene} authors Unreal cutscenes through tools and checks call trajectories against an API-derived dependency graph, which is trajectory validity rather than world-state validity, with no baseline comparison or ablation. The synthetic-data tradition, from Playing for Data~\cite{richter2016playing} to BEDLAM~\cite{black2023bedlam}, VirtualHome~\cite{puig2018virtualhome}, and Action Genome~\cite{ji2020actiongenome}, proved long ago that simulation buys perfect annotation at scale. In every one of those platforms, a researcher authored the specification, by hand or by code.

Our world is neither a menu nor a script library. Earlier text-to-GEST prototypes~\cite{masala2023gest} demonstrated the translation in principle, without targeting executability. To author an executable GEST, an agent must keep a story coherent and simulator-valid at the same time, across the dozens of events of a multi-actor story (7--65 per scenario in the corpus~\cite{cudlenco2026gtasagroundtruthannotations}): action chains, object lifecycles, capacities of points of interest (POIs), cycle-free temporal constraints. LLMs are excellent at the first requirement and fail at the second. We built the obvious system first, a staged pipeline driving a flagship LLM through six validated stages, and watched it fail 50 times out of 50 (Section~\ref{sec:staged}).

The architecture that works separates concerns: the LLM decides what should happen, a programmatic state backend decides what is allowed to happen, so every specification the agents emit is valid by construction. Section~\ref{sec:architecture} describes the environment and the agents that author through it; Section~\ref{sec:evaluation} measures executability and reconstruction fidelity.

Our contributions are:
\begin{itemize}
    \item \textbf{A constraint-enforcing authoring environment} (more than 30 validated tools over a stateful backend with transactional chains, capacity tracking, object lifecycles, cycle detection, and synchronized interactions) and a hierarchical Director / Scene Builder architecture that authors multi-scene stories through it, raising executability from 0 of 50 for a flagship model in the standard staged design (Section~\ref{sec:staged}) to 20 of 25 for a budget model.
    \item \textbf{A reconstruction-fidelity evaluation} against plan-level ground truth, a measurement unique to this setting: the seed text derives from a source graph, and the agent reconstructs its events at 0.83 F1 (0.55 for a random baseline), with the residual dominated by information the text channel drops.
    \item \textbf{The first system, to our knowledge, to exercise the full expressive capacity of the GEST formalism} (hierarchy, synchronized interactions, semantic and logical structure), closing the loop from story, to an explainable scene representation, to video with ground truth, and back to text.
\end{itemize}

\section{The Staged Pipeline and Why It Fails}
\label{sec:staged}

Before the agentic architecture, we built a staged pipeline: a LangGraph workflow that drives GPT-5, the flagship tier of its provider at the time, through a fixed graph of six specialized stages. \emph{Concept} drafts an abstract hierarchical GEST with parent and leaf scenes and actor archetypes; \emph{Casting} assigns skins from the engine's catalog; \emph{Episode Placement} maps scenes to valid simulation episodes; \emph{Setup} plans off-camera preparation and backstage positioning; \emph{Screenplay} translates the abstract narrative into concrete action sequences; \emph{Scene Detail} expands each leaf scene into a complete executable GEST. We ran the pipeline on 50 stories, each stage prompting for structured output and parsing it, re-prompting up to three times on failure (an API error or invalid GEST JSON). We obtained coherent narratives with logical structure and character motivation, and \textbf{zero} executed. 45 of the 50 produced a complete graph; the remaining five stalled in the detailing stages after concept and casting, never emitting one. The failure is mechanical, not accidental: the engine can place none of the 45, because their scenes demand what no episode of the world can host.

The failure modes compound across stages: \emph{near-miss action ids} (``LookAtWatch'' for the world's ``LookAtTheWatch''; 8 of the 45 completed graphs), \emph{state-rule violations} (an actor seated again without ever standing up; 21 of 45), \emph{impossible placements} (laptop action chains demanded in a living room, where the world places no laptop), \emph{object lifecycle violations} (putting down objects never picked up; 5 of 45), and \emph{rule neglect}: each stage receives a carefully formalized, curated slice of the 14{,}000-line capability registry (i.e., the Scene Detail prompt has roughly 3{,}800 lines of rules for action chains, temporal relations, example reference graphs, and episode data), and the model still violates rules stated verbatim in its context. These map onto exactly the two properties that make this world hard to author for: the registry is never fully in context, and validity is state-conditional. Each stage makes locally reasonable decisions, but accumulated state (who is where, what they hold, which POIs are occupied, which constraints are in effect) drifts from validity as the GEST grows. And the errors cannot be repaired afterwards: no fixed normalization covers non-deterministic hallucinations, and fixing one violation invalidates the state that later events depend on, so correction backtracks at superlinear cost. Validity must be enforced where the specification is written, not checked after it is finished. The lesson: \emph{the LLM should decide what should happen; a programmatic backend should decide what is allowed to happen.}

\section{The Agentic Architecture}
\label{sec:architecture}

We built the agentic system on LangGraph's deepagents: \emph{reactive} agents~\cite{yao2022react} organized through hierarchical subagent delegation, all driven by Claude Haiku 4.5, its provider's budget tier at the time. The Director plans the story, and the Scene Builder implements it as GEST, scene by scene. Every authoring session opens with \textbf{the Director Agent}, which receives an optional seed text and a generation configuration and works in four phases. \emph{Exploration:} read-only tools scope down the 14{,}000-line capability registry (episodes, regions, POI action chains, and character skins with visual descriptions), paginating through it in smaller chunks. The explore-before-plan pattern prevents hallucination: an action enters a plan only after a tool call has confirmed it exists. \emph{Casting:} the Director creates the story and its actors (name, gender, skin, starting region). \emph{Scene building:} the Director processes scenes sequentially. For each, it selects episode, region, and participants, delegates to the Scene Builder with a natural-language brief, and moves actors between regions afterwards. \emph{Finalization:} the Director links scene boundaries with cross-scene temporal constraints.

\textbf{The Scene Builder Subagent} receives an isolated context (scene, episode, region, actors, and the brief) and never sees the rest of the story. It constructs events through a round-based state machine. A round opens with the current state of every actor (posture, held objects, location). For each actor, the subagent starts a chain at a POI and extends it among the valid continuations the backend returns. For example, an actor sitting down at a desk that has a laptop occupies that seat for everyone else, and may then open the laptop or stand up. Opening the laptop adds typing on it or closing it as continuations. Synchronized two-actor interactions, camera control, and cross-actor temporal dependencies are separate validated calls. Closing the round commits cross-actor ordering. Chains commit atomically: events accumulate in a buffer, only an explicit commit writes them into the GEST, and failed explorations leave no trace.

\begin{figure}[t]
\centering
\includegraphics[width=\textwidth]{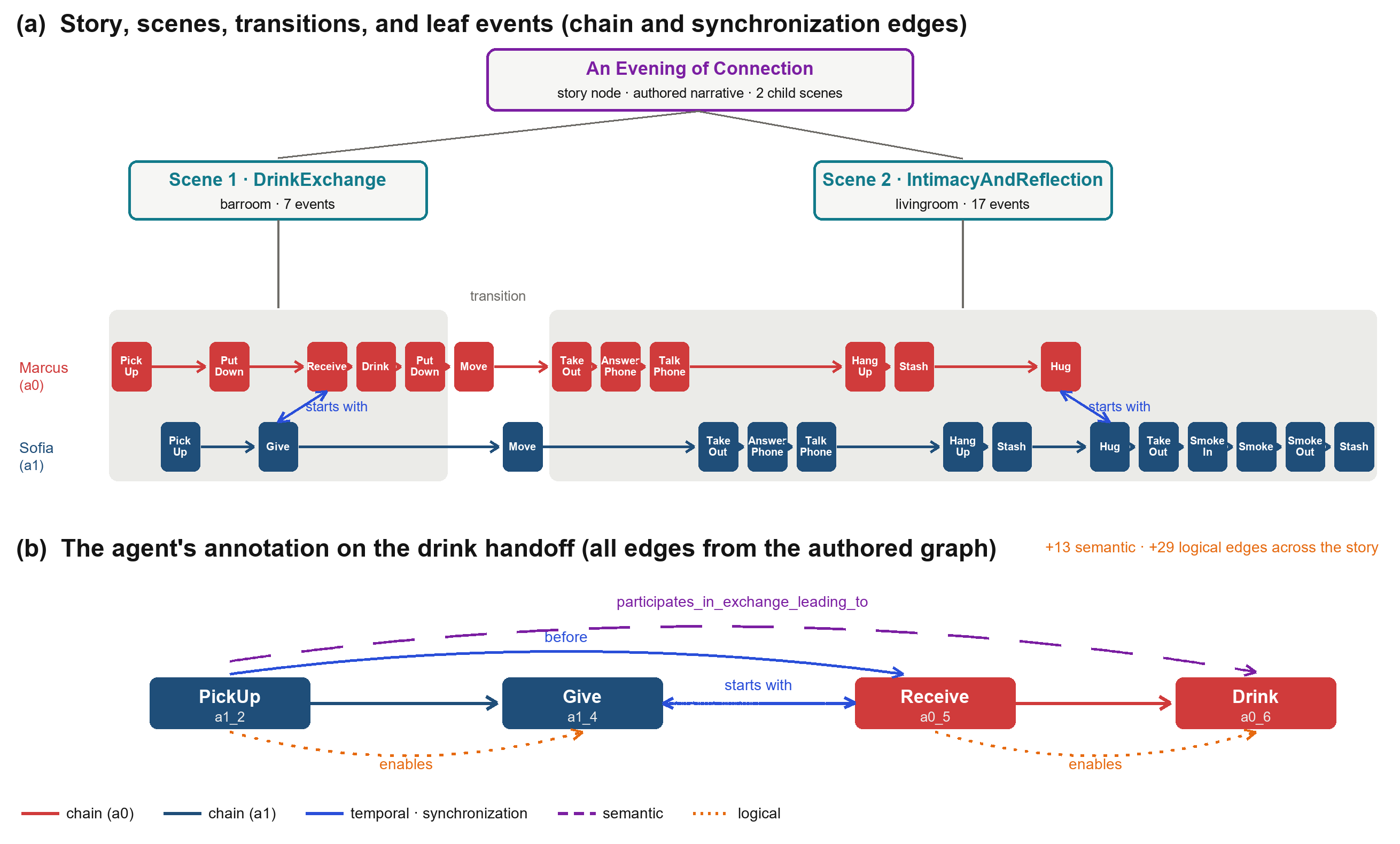}
\caption{\textbf{Anatomy of an agent-authored GEST} (seeded story \emph{An Evening of Connection}). (a)~The Director produces the story node and decomposes it into scenes. The Scene Builder writes every node inside each scene, including synchronized pairs (give\,$\leftrightarrow$\,receive). The Director glues the scenes with \texttt{Move} events and temporal constraints. (b)~The drink-handoff close-up: the Relation Subagents add the logical and semantic edges last (13 semantic, 29 logical in this story).}
\label{fig:gest_anatomy}
\end{figure}

After each scene and after final assembly, the Director optionally delegates to two further subagents: a \textbf{Logical Relations Agent} adds causal and dependency edges (\texttt{causes}, \texttt{enables}, \texttt{prevents}, \texttt{requires}), and a \textbf{Semantic Relations Agent} adds narrative-coherence edges with free-text types (\texttt{observes}, \texttt{interrupts}, \texttt{motivates}). These edges are optional for execution, but they populate the logical and semantic edge types defined in the GEST formalism that procedural generation never produces; to our knowledge this is \textbf{the first system to exercise the full expressive capacity of the representation} in one pipeline.

Every operation of every agent passes through \textbf{the state backend} (Fig.~\ref{fig:hero}, top): the engine's procedural random generator, extended with a delegation interface that exposes its operations as more than 30 validated tools. The tools implement a transaction-based state machine over four states (\texttt{IDLE}, \texttt{STORY\_CREATED}, \texttt{IN\_SCENE}, \texttt{IN\_ROUND}). Behind them, the backend holds the world state and reshapes what they offer after every choice. It enforces POI capacity for exclusive-use objects. It walks the dependency graph before accepting any \texttt{before} relation and rejects edges that would create a cycle. It coordinates two-actor interactions: co-location, posture, and an explicit receiver for \texttt{Give} with an automatically synchronized receiving event. It locks atomic object sequences (\eg{} \texttt{TakeOut}$\to$\texttt{Use}$\to$\texttt{Stash}) until completion. Rejections return explanatory errors the agent uses to re-plan: the environment's dynamics, seen from the agent's side.

The LLM decides what should happen but chooses from what the state affords, never needing the whole world in memory. Nothing touches the GEST except through the backend, so authoring inherits the validity guarantee of the random generator it is built on. Rendering is offline, but authoring is an agent-environment loop under partial observability, the interaction problem embodied agents face. The division is also cheap: with the constraints in the tools, a budget model suffices, and a complete multi-scene story costs approximately \$0.25 in API calls.
\section{Evaluation}
\label{sec:evaluation}

\begin{table}[t]
\centering
\caption{Executability: authoring attempts whose GEST simulates end to end.}
\label{tab:gen_stats}
\begin{tabular}{@{}lccc@{}}
\toprule
Approach & Attempted & Successful & Rate \\
\midrule
Staged workflow & 50 & 0 & 0\% \\
Tool-constrained agent & 25 & \textbf{20} & \textbf{80\%} \\
\bottomrule
\end{tabular}
\end{table}

\subsection{Corpus and Protocol}

We evaluate on GTASA~\cite{cudlenco2026gtasagroundtruthannotations}, the corpus produced with the GEST-Engine~\cite{cudlenco2026gestengineeventgraphssynthetic}. Its stories are diverse in scene type (classroom, gym, garden, house, mixed) and actor count (2--6). Each comes with its GEST, its execution video, and a textual description. To obtain the description, the graph is transformed into a proto-language, and an LLM then refines it into natural language without altering actors, events, or order of actions. The corpus's human studies found the engine's videos physically valid in 69\% of cases (VEO~3.1: 18\%, WAN~2.2: 13\%) and a closer match to the GEST-derived description, the same text the neural generators were conditioned on (semantic match 4.09/5 versus 2.50 and 1.75). Our agents drive the same executor, so our videos inherit those results. What is new is the authoring: the system turns free text into video, closing the loop from text, to graph, to video, and back to text.

We select 25 corpus stories and hand each description to the Director as seed text. Across the staged baseline and the seeded runs, we evaluate 75 authoring attempts end to end, at the scale of the field: FilmAgent~\cite{xu2025filmagent} evaluates on 15 story ideas, the Cutscene Agent~\cite{he2026cutscene} on 65 scenarios. Seeding from the corpus, and not from hand-authored stories that carry no source graph to score against, makes two measurements possible. First, executability: does the agent author a new GEST that the engine accepts and simulates, an attempt counts as successful only then. Second, reconstruction: how similar is the authored graph to the source graph behind the text. We report both in full below.

\subsection{Executability}

We define executability as follows: a GEST is useful only if it produces a video, which it does only if it is valid and the engine executes it without runtime errors. No published system authors executable GESTs, so there is no external baseline. The staged workflow fills that role: it follows the standard staged LLM planning design~\cite{lin2023videodirectorgpt,fei2024dysen,hu2024storyagent} and drives the flagship GPT-5 with per-stage structured validation. It produced narratively coherent stories, and none executed (Section~\ref{sec:staged} explains why). The tool-constrained agent, driving the budget Claude Haiku 4.5, executed 20 of 25 (80\%). Since the winning side runs the weaker model, the gap cannot be a model effect. What changed is where the constraints live: in the prompt as instructions, or in the tool layer as checks no output can bypass. The agent's five failures are not constraint violations either, since its graphs are valid by construction: all five were picked up by the engine, then crashed or stalled at runtime until their simulation retries ran out, the same engine-side failure modes the corpus pipeline reports for the procedural generator's valid-by-construction graphs~\cite{cudlenco2026gestengineeventgraphssynthetic}. Table~\ref{tab:gen_stats} summarizes the results.

\subsection{Reconstruction Fidelity}
\label{sec:reconstruction}

Every seed text was derived from a source graph $G$, which makes $G$ a plan-level ground truth for the agent's output $\hat{G}$. We can therefore measure how much of the story survives the round trip graph $\to$ text $\to$ agent $\to$ graph. To our knowledge, \textbf{no other system in this space has a plan-level ground truth to reconstruct against}.

We score reconstruction over the 20 attempts that executed. We list the events of each graph and match them one to one, at three increasingly strict levels: by action alone; adding the types of the participating entities; adding the location. Two events match only if these descriptions are identical, and each event can match at most once. Matched events count as true positives: precision penalizes the events the agent added, recall the events it missed, and we report their F1. Sequential structure is scored the same way over per-actor bigrams, consecutive action pairs in each actor's chain. As a random baseline we score each source plan against three \emph{random graphs}, randomly selected corpus scenarios matched in scene type and actor count (60 comparisons over the 20 pairs).

\begin{table}[t]
\centering
\caption{Reconstruction fidelity over 20 seeded pairs (mean$\pm$std F1). Each column scores a graph against the source plan whose text seeded the agent: the \emph{agentic graph}, authored from that text alone, and a \emph{random graph}, a randomly selected corpus scenario matched in scene type and actor count (3 per source, 60 comparisons in total).}
\label{tab:reconstruction}
\begin{tabular}{@{}lcc@{}}
\toprule
Level & Agentic graph & Random graph \\
\midrule
Action & \textbf{0.85}$\pm$0.15 & 0.61$\pm$0.19 \\
Action + entities & \textbf{0.83}$\pm$0.18 & 0.55$\pm$0.21 \\
Action + entities + location & \textbf{0.63}$\pm$0.39 & 0.35$\pm$0.30 \\
Sequential structure (bigrams) & \textbf{0.77}$\pm$0.21 & 0.43$\pm$0.21 \\
\bottomrule
\end{tabular}
\end{table}

Table~\ref{tab:reconstruction} reports the scores, and Fig.~\ref{fig:gest_anatomy} shows one reconstructed story in full: every event of the source plan reappears (recall 1.0; F1 0.70, as the agent also adds a phone interlude and a second scene), and the drink-handoff motif survives the round trip with its synchronization intact, grounded in a different room because the seed text never names one. The agent beats its per-pair baseline mean in 19 of 20 pairs on events and 20 of 20 on sequential structure (sign test, $p < 10^{-4}$). Four pairs reconstruct perfectly (F1 $=$ 1.0 at every level), verified as genuine rebuilds with different node identities rather than copies, so the ceiling is attainable when the text preserves the information. The location level explains the residual: the agent grounds the story in the source region precisely when the seed text names it, and the five classroom texts, which never do, are exactly the five pairs scoring 0 on location. What the language channel drops, no agent can recover; what it keeps, the agent reconstructs.

\section{Conclusion}
\label{sec:conclusion}

We presented an agentic system that authors executable specifications for a living world: from free text or exploration, a Director and a Scene Builder construct multi-scene, multi-actor GESTs through a constraint-enforcing tool layer, and the engine executes them into videos with exact per-frame ground truth; three conclusions follow.

\textbf{First}, the architecture is the result: moving constraint enforcement from the model into more than 30 validated tools over a transactional state backend raises executability from 0 of 50 (a flagship model in a staged workflow) to 20 of 25 for a budget model, and makes invalid specifications impossible rather than improbable.

\textbf{Second}, reconstruction against plan-level ground truth, a measurement unique to this setting, reaches 0.83 event F1 over a 0.55 random baseline, with the residual exactly where the text drops information.

\textbf{Third}, to our knowledge the system is the first to exercise the full expressive capacity of the GEST formalism (hierarchy, synchronized interactions, semantic and logical structure) in a single pipeline, \textbf{closing the loop from story, to an explainable scene representation, to video with ground truth, and back to text} through the GEST-derived description.

\textbf{Limitations and future work.} We will continue working toward filmmaker-like workflows that use everything the engine offers: posing actors off-camera and shooting only the moments that matter, as montage; authoring stories with more scenes; and giving the Director images of those scenes as input, toward fuller, more complex narratives.

\textbf{Code and data.} The authoring system is available at \url{https://github.com/ncudlenco/multiagent_story_system/releases/tag/v1.1-eccv2026} and the GEST-Engine at \url{https://github.com/ncudlenco/mta-sim/releases/tag/v1.0-iclr2026}, both tagged at the state used for the results reported here. The GTASA corpus is available at \url{https://huggingface.co/datasets/nnc-001/gtasa-01}.

\subsubsection*{Acknowledgements.}
This work is supported in part by B\"uchi Labortechnik AG and by the projects
``Romanian Hub for Artificial Intelligence --- HRIA'', Smart Growth,
Digitization and Financial Instruments Program, 2021--2027 (MySMIS no.~334906)
and ``European Lighthouse of AI for Sustainability --- ELIAS'', Horizon Europe program (Grant No.~101120237).

\bibliographystyle{splncs04}
\bibliography{main}

\end{document}